\documentclass[letterpaper, 10 pt, conference]{ieeeconf}  

\usepackage{caption}
\usepackage{subcaption}
\makeatletter

\renewcommand{\p@subtable}{\thetable-}
\makeatother
\usepackage{amsmath,amsfonts}
\usepackage{algorithmic}
\usepackage{array}
\usepackage[caption=false,font=normalsize,labelfont=sf,textfont=sf]{subfig}
\usepackage{textcomp}
\usepackage{stfloats}
\usepackage{url}
\usepackage{verbatim}
\usepackage{graphicx}
\usepackage{booktabs}
\usepackage{stmaryrd}
\usepackage{multirow}
\usepackage{pifont}
\usepackage{makecell}
\usepackage{float}

\def\BibTeX{{\rm B\kern-.05em{\sc i\kern-.025em b}\kern-.08em
    T\kern-.1667em\lower.7ex\hbox{E}\kern-.125emX}}
\usepackage{balance}
\usepackage[compress]{cite}
\usepackage[dvipsnames]{xcolor}

\newcommand{\method}{SAFe}
\newcommand{\cmark}{\ding{51}}
\newcommand{\xmark}{\ding{55}}

\usepackage{isomath}
\usepackage{bm}
\DeclareMathAlphabet{\mathsfit}{\encodingdefault}{\sfdefault}{m}{sl}
\SetMathAlphabet{\mathsfit}{bold}{\encodingdefault}{\sfdefault}{bx}{n}
\newcommand{\tens}[1]{\bm{\mathsfit{#1}}}

\IEEEoverridecommandlockouts                              

\title{\LARGE \bf
SAFe: Segment-guided Aggregation of Feature Densities for Anomaly-aware Segmentation
}

\author{Anja Delić, Jurica Runtas, Marin Oršić, Ivan Marković, Ivan Petrović \footnote{}\\
University of Zagreb Faculty of Electrical Engineering and Computing\\
Laboratory for Autonomous Systems and Mobile Robotics \\
Zagreb, Croatia \\
\tt\small name.surname@fer.hr
}

\begin{document}

\maketitle
\thispagestyle{empty}
\pagestyle{empty}

\begin{abstract}
Visual segmentation systems encounter objects outside their training distribution during real-world deployment,
hindering reliable autonomous systems that depend on scene parsing in the perception stage.
Many recent methods address this by using self-supervised foundation models to train
density estimators that yield low likelihood in anomalous image regions.
Although promising, these methods suffer from poor feature semantics or they lack spatial
consistency, both of which undermine critical downstream decisions.
We address this problem with~\method, a generative method based on class-conditional density
estimation over self-supervised representations.
\method~trains lightweight normalizing flows that produce class-conditional normalized
likelihood estimates over frozen DINOv3 features.
We combine density estimates from transformer features with density scores over
multi-scale convolutional features to capture both global semantics and local detail.
We introduce a method-agnostic post-processing step based on SAM3 that connects
per-location likelihoods into spatially coherent segments while suppressing false
positives, and enables instance-level anomaly detection without retraining.
The post processing further distinguishes novel categories among anomalous objects by a similarity-based agglomerative clustering scheme.
\method~sets a new state of the art on the PANIC, OoDIS, SMIYC ObstacleTrack with strong performance on the ISSU benchmark.
\end{abstract}

\section{Introduction}
Modern autonomous systems increasingly rely on learning-based visual perception,
yet unstructured deployment environments inevitably contain visual concepts outside the training distribution.
If the perception component does not address novel concepts during inference,
missclassification into familiar concepts propagates to planning and control,
potentially leading to critical decision making failures~\cite{latombe1989motion}.
This is particullary problematic in scene segmentation which commits to classification in every pixel~\cite{chan2021iccv}.
Anomaly-aware segmentation addresses this failure mode by complementing each classification with an anomaly score.
This promotes downstream safety mechanisms such as path planning with collision-avoidance~\cite{wellhausen2020ral}, fallback planning and control~\cite{sinha2024rss}, or post-hoc behavior analysis.

\begin{figure}[h]
    \centering
    \includegraphics[width=\linewidth]{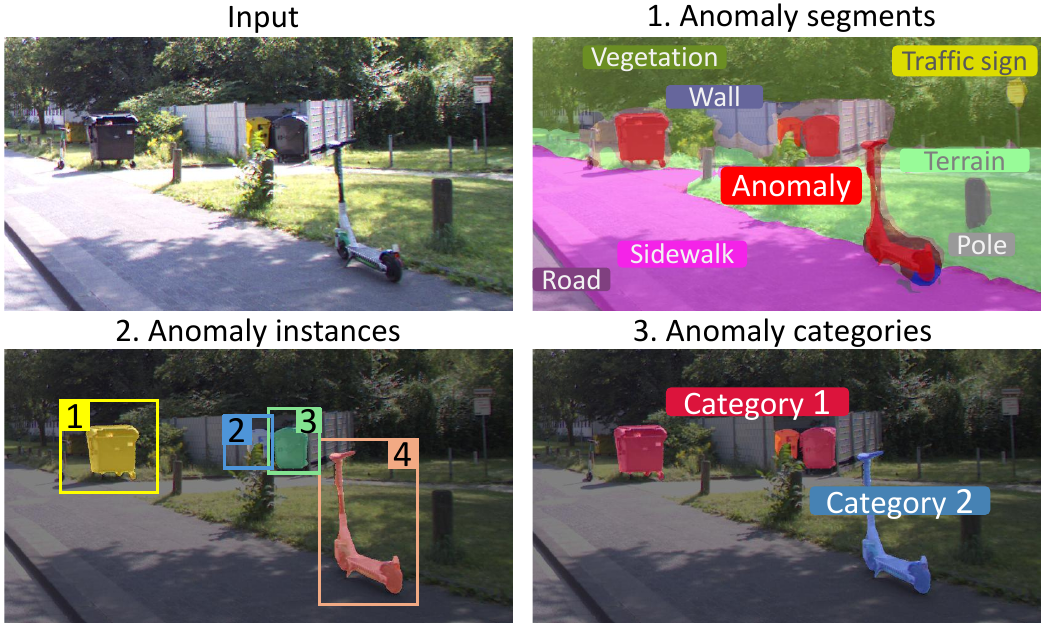}
    \caption{\method~reasons about anomalies at three levels: (1) detects anomaly segments and classifies segments in the inlier region, (2) distinguishes instances in the anomalous region, and (3) groups anomalous instances into novel categories.}
    \label{fig:tasks}
\end{figure}

Recent work shows that self-supervised foundation models surpass domain-specific backbones for anomaly segmentation,
since large-scale pre-training yields more general representations~\cite{vojivr2024eccv, zheng2025icm, lee2026wacv}.
Furthermore, generative modeling of foundation features enables principled scoring via likelihood estimation,
with semantic labels incorporated through class-conditional density estimation or contrastive training~\cite{liang2022neurips,lee2026wacv}.
Normalizing flows are particularly compelling as they model exact feature likelihood under the inlier distribution~\cite{rezende2015icml},
yet prior methods estimate density over features extracted using either supervised backbones~\cite{blum2021ijcv,gudovskiy2022wacv, gudovskiy2023uai, liang2022neurips}
or self-supervised transformers that are semantically strong but spatially less coherent than convolutional features~\cite{vojivr2024eccv,lee2026wacv,simeoni2025arxiv}.
To increase spatial consistency of anomaly segmentation, leading methods reason at the segment level using
either Mask2Former~\cite{delic2024bmvc, nayal2023iccv, grcic2023cvprw} or Segment Anything~\cite{carion2025arxiv, ravi2024arxiv, kirillov2023iccv} models.
However, existing approaches do not jointly exploit the complementary semantic and local properties of foundation representations while enforcing segment-level spatial consistency during anomaly estimation.

We bridge this gap with~\method~(\textbf{S}egment-guided \textbf{A}ggregation of \textbf{F}eature D\textbf{e}nsities).
\method~trains multiple class-conditional normalizing flows that independently estimate per-position density in the frozen foundation model feature map.
Each normalizing flow operates on a different feature map from DINOv3 backbones, therefore we combine the density estimates into a compound anomaly score.
Concretely, we find that Vision Transformer (ViT) improves global semantics, whereas ConvNeXt retains local detail, and the combined anomaly estimates perform the best.
We bridge dense per-location anomaly estimates and object-level predictions with a method-agnostic aggregation refinement framework based on SAM3 that also improves existing methods.
Prompted with inlier class descriptions, this refinement suppresses false positives within generated segments.
The refinement also supports instance-level anomaly estimates as it aggregates per-location scores into mask-level predictions.
Because anomalous regions contain objects of previuosly unseen concepts, we identify novel semantic categories among predicted instances via clustering in the feature space.
\method~thus detects anomalies and reasons at the levels of segments, instances, and categories, as illustrated in Fig.~\ref{fig:tasks}.

In summary, our contributions are:
(i)~the first anomaly-aware segmentation method to perform exact class-conditional likelihood estimation over frozen foundation features,
(ii)~a normalization and fusion scheme that merges complementary density estimates into a composite anomaly score,
(iii)~a method-agnostic, segment-guided refinement framework that also improves existing anomaly-segmentation methods and enables instance-level recognition without retraining, and
(iv)~anomaly category discovery based on clustering of feature similarity agnostic to the anomaly category set cardinality.
\method~attains state-of-the-art results on on the PANIC panoptic tasks~\cite{sodano2024arxiv}, OoDIS instance detection~\cite{nekrasov2025icra}, SMIYC ObstacleTrack anomaly segmentation~\cite{chan2021neuripsdb} with strong performance in anomaly and open-set segmentation on the ISSU benchmark ~\cite{laskar2025cvpr}.

\section{Related work}
\label{sec:related_work}

\noindent \textbf{Anomaly detection over foundation backbones.}
Early approaches to anomaly segmentation rely on prediction uncertainty of convolutional models~\cite{chan2021iccv}, later superseded by mask transformers~\cite{cheng2022cvpr, grcic2023cvprw, rai2023iccv, ackermann2023bmvc, nayal2023iccv, delic2024bmvc}.
Recent approaches instead leverage large-scale foundation models such as self-supervised vision transformers~\cite{caron2021iccv}, promptable segmentation models~\cite{kirillov2023iccv}, and vision-language models~\cite{yu2023neurips} to obtain representations that generalize beyond the in-distribution taxonomy.
PixOOD~\cite{vojivr2024eccv} leverages a frozen DINOv2 ViT~\cite{oquab2023arxiv} encoder alone, but its patch tokenization limits localization of small anomalies.
VL4AD~\cite{zhong2024vl4ad} scores anomalies via cosine-similarity between visual and CLIP text embeddings, exploiting textual OOD prompts, but hinges on a hand-crafted concept dictionary.
SOTA~\cite{zheng2025icm} fine-tunes SAM~\cite{kirillov2023iccv} via LoRA to refine anomaly scores from off-the-shelf detectors.
Unlike prior work, we rely on the most recent DINOv3~\cite{simeoni2025arxiv} and SAM3~\cite{carion2025arxiv} without any fine-tuning, leveraging DINOv3's transformer and convolutional variants and refining pixel-level scores with SAM3 segments prompted by point locations and in-distribution class names.

\noindent \textbf{Generative models for anomaly detection.}
Early approaches model inlier feature density with normalizing flows~\cite{blum2021ijcv, gudovskiy2023uai}, Gaussian mixtures~\cite{liang2022neurips}, or hybrid formulations~\cite{grcic2024tpami}, but underperformed discriminative alternatives.
Recent work instead builds on frozen DINOv2~\cite{oquab2023arxiv} features: FlowCLAS~\cite{lee2026wacv} trains a normalizing flow with contrastive learning, while UEM~\cite{nayal2025ijcv} and ULRE~\cite{holle2025iccv} frame detection as a likelihood-ratio test between learned outlier and inlier distributions, the latter adding evidential uncertainty over the ratio.
In contrast to \method~, all three depend on a proxy outlier distribution from outlier exposure~\cite{hendrycks2018arxiv}, whose quality and coverage bound performance.

\noindent \textbf{Anomaly instance detection.}
Anomaly instances are typically separated either from an already computed anomaly score, or by training a dedicated instance branch.
Prior2Former~\cite{schmidt2025iccv} clusters mask embeddings within a thresholded uncertainty region, while UGainS~\cite{nekrasov2023gcpr} samples points from a thresholded uncertainty map to prompt SAM~\cite{kirillov2023iccv}.
Mask2Anomaly~\cite{rai2023iccv} extracts connected components in the remaining background region, and Con$2$MAV~\cite{sodano2024arxiv} trains a dedicated instance decoder jointly.
Unlike others, \method~is implicitly instance-aware and takes prior obtained SAM3 segments as anomaly instances.

\noindent \textbf{Anomaly category discovery.}
Distinguishing categories of detected anomalous objects remains underexplored.
Existing approaches mostly leverage DBSCAN~\cite{schmidt2025iccv, gasperini2023iccv} or HDBSCAN~\cite{sodano2024arxiv} in feature space, since the number of clusters is not known a priori.
We show that these methods tend to collapse or over-segment, and instead propose agglomerative clustering with early stopping by a similarity criterion.

\section{The Proposed Method}
\label{sec:method}
\begin{figure*}
\vspace{8pt}
    \centering
    \includegraphics[width=\textwidth]{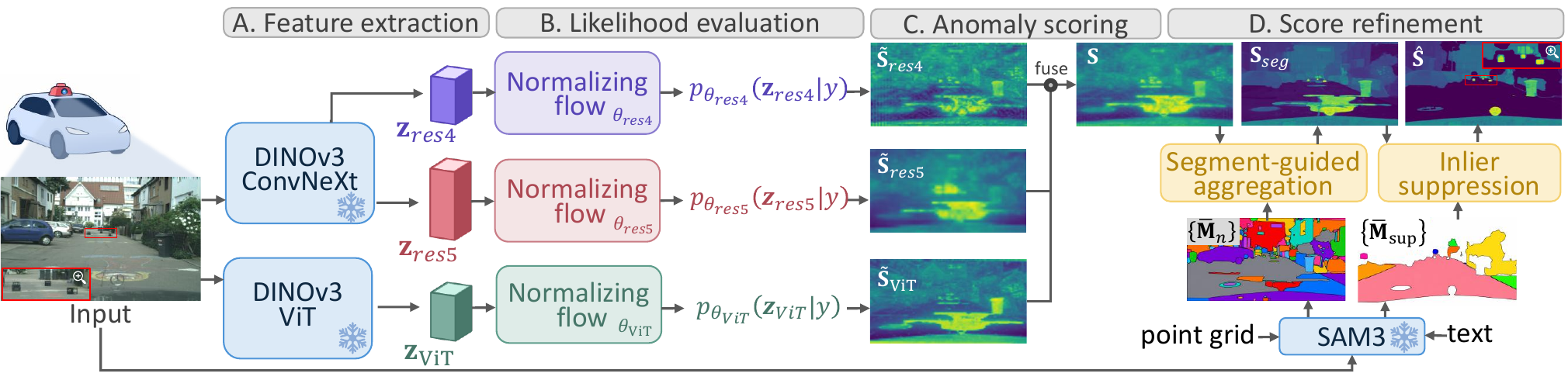}
    \caption{The proposed anomaly scoring pipeline: (A) per-location feature extraction using frozen DINOv3 trained backbones, (B) per-stage feature scoring with class-conditional normalizing flows, (C) normalization and fusion of per-stage likelihoods (D) score aggregation by SAM3 point-prompted masks and suppression at locations indicated by text-prompted masks. \method~successfully raises the anomaly score for road obstacles and suppresses false high scores on the drawing on the road. }
    \label{fig:method}
\end{figure*}

\method~frames anomaly segmentation as class-conditional density estimation 
in the feature space of two frozen DINOv3-pretrained encoders, 
ViT and a ConvNeXt, 
with predictions spatially grounded by class-agnostic segments from 
SAM3~\cite{simeoni2025arxiv, dosovitskiy21iclr, carion2025arxiv, liu22cvpr}.
The anomaly scoring framework, as illustrated in Fig.~\ref{fig:method}, consists of four components: 
(A) dense multi-scale feature extraction using the two backbones, 
(B) class-conditional density modeling with normalizing flows, 
(C) anomaly scoring through likelihood evaluation, 
and (D) segment-guided anomaly score aggregation and inlier suppression.
The resulting anomaly score further supports instance-level anomaly detection (Sec.~\ref{sec:instances}), and anomaly category discovery (Sec.~\ref{sec:discovery}).

\subsection{Multi-backbone dense feature extraction}
\label{sec:feature_extractor}

We process input images in parallel using frozen DINOv3 ViT and ConvNeXt backbones.
We freeze both backbones to preserve their generalized 
representations and keep the trainable parameter count low.
We retain the feature maps corresponding to
the final ViT output
and 
the \textit{res4} and \textit{res5} stages of ConvNeXt:
\begin{equation}
\label{eq:feature}
    \tens{F}_l \in \mathbb{R}^{H_l \times W_l \times C_l}, \quad l \in \{\mathrm{ViT, res4, res5}\}
\end{equation}
where
$H_l$, $W_l$, and $C_l$ 
denote feature map height, width and channels, respectively.
Precisely, 
$\tens{F}_{\mathrm{ViT}}$ and $\tens{F}_{\mathrm{res4}}$ operate at 
stride 16, while $\tens{F}_{\mathrm{res5}}$ is spatially coarser at stride 32.
The two backbones provide complementary representations: the global self-attention in ViT produces discriminative semantic features well-suited to detecting large anomalous objects,
while
the convolutional feature pyramid preserves local interactions that a standalone ViT stream would otherwise miss.

\subsection{Class-conditional normalizing flows} 
\label{sec:normalizing_flows}

We treat each spatial location in a feature map $\tens{F}_l$ as an independent sample $\mathbf{z}_l^{(h,w)} \in \mathbb{R}^{C_l}$ from location $(h,w)$.
This allows us to use 1D normalizing flows which are computationally 
efficient and enable application to arbitrary image resolutions, unlike 2D flows that model full spatial dependencies and 
are tied to a fixed spatial extent~\cite{lee2026wacv}. 
We subsequently recover the lost spatial correlation 
through segment-guided aggregation with SAM3
(Sec.~\ref{sec:sam_refinement}).

Foundation model features encode rich class-discriminative information even without task-specific supervision~\cite{simeoni2025arxiv}. Thus, we condition the normalizing flow on class information to capture per-class statistics.
Formally, for a feature vector 
$\mathbf{z} \in \mathbb{R}^{C_l}$ extracted at a spatial location at the feature stage $l$ (Eq.~\ref{eq:feature}), 
we learn an invertible transformation 
$\mathbf{u} = f_{\theta_l}(\mathbf{z} \mid y)$ 
that maps $\mathbf{z}$ 
to a latent variable $\mathbf{u}$ and is
conditioned on the class label $y$.
For every feature stage $l$ we instantiate a separate model, allowing the models to learn stage-specific feature statistics.

\begin{figure}[t]
    \centering
    \includegraphics[width=\linewidth]{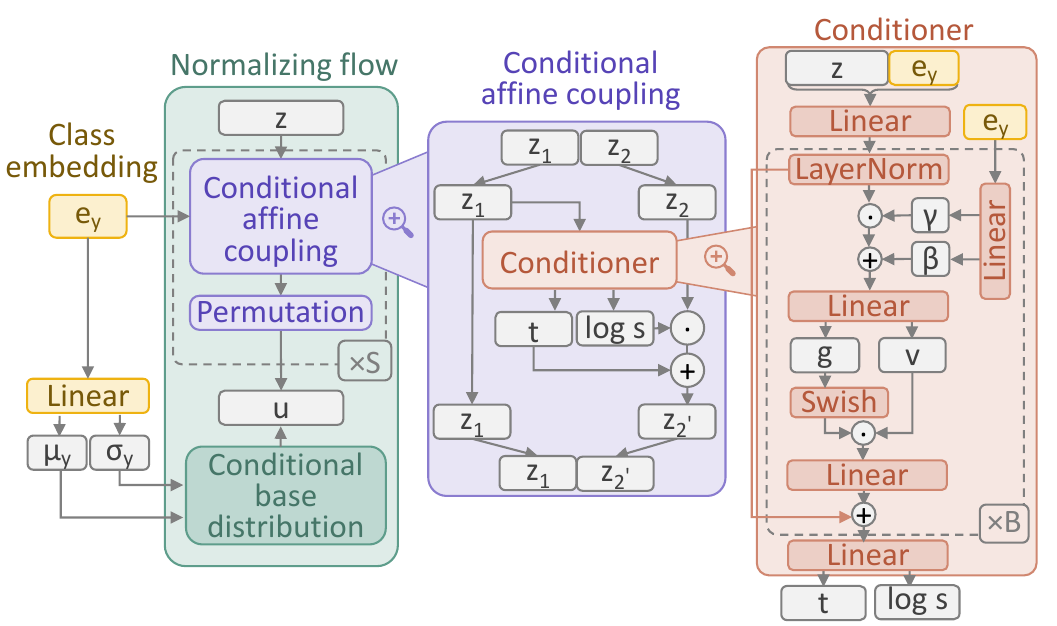}
    \caption{Architecture of the normalizing flow (green), conditional affine coupling transformations (purple), conditioner module (red) and the class-conditioning mechanism (yellow).}
    \label{fig:nf}
\end{figure}

We design the normalizing flow $f_{\theta_l}$, illustrated in Fig.~\ref{fig:nf}, as a sequence of $S$ conditional affine 
coupling transformations with channel-wise partitioning, interleaved with fixed 
channel-wise permutations~\cite{kingma2018neurips}.
For every known class we jointly train a class embedding $\mathbf{e}_y\in \mathbb{R}^{d_e}$ used to
forward class cues to the base distribution and the coupling transformations (Fig.~\ref{fig:nf} in yellow).
The base distribution consists of separate Gaussians for every known class
$p_{\mathbf{U}\mid\mathbf{Y}}(\mathbf{u}|y)
= \mathcal{N}\!\bigl(\mathbf{u}\mid\bm{\mu}_y,\, \operatorname{diag}(\bm{\sigma}^2_y)\bigr).$
We express $\bm{\mu}_y$ and $\bm{\sigma}_y$ as trainable embedding lookup using class index $y$~\cite{winkler2019arxiv}.
This allows the latent space to adapt to the classes, rather than forcing all classes to map to the same Gaussian.
The log-likelihood follows from the change-of-variables formula \cite{kingma2018neurips}:
\begin{equation}
\log p_\theta(\mathbf{z} \mid y) = \log p_{\mathbf{U}\mid\mathbf{Y}}(\mathbf{u} \mid y) + \sum_{j=1}^{S} \sum_{i} \log s_{j,i},
\end{equation}
where $s_{j,i}$ is the scale output of the $i$-th channel in the $j$-th coupling step.

The conditioner module of a coupling transformation (Fig.~\ref{fig:nf} in red) first concatenates the class embedding $\mathbf{e}_y$ to the conditioner input.
The result is then linearly projected and normalized.
We apply channel-wise affine modulation (FiLM) \cite{perez2018aaai} to this normalized activation to boost class-specific statistics.
The class-conditioned activations then pass through a gating mechanism that acts as a channel-wise nonlinear feature selector \cite{dauphin2017icml, shazeer2020arxiv}. We use the Swish \cite{ramachandran2017arxiv} activation $\mathrm{Swish}(x)=x \cdot \sigma(x)$ as the gate nonlinearity, since it avoids the hard cutoff of ReLU-based gating.
Finally, we stack $B$ such blocks of normalization, FiLM modulation, and gating each with a residual connection.
Our empirical results (Sec.\ref{sec:discussion-class-cond}) show that conditioning at all these axes contributes to overall performance.

We extract features from training images offline. We train each flow to maximize the log-likelihood given cached feature vectors and corresponding labels $y$.
During training, $y$ is provided by the Cityscapes ground-truth semantic labels and we consider only features corresponding to image patches with more than 90\% of
one label~\cite{vojivr2024eccv}. 
During inference, we marginalize the likelihood through model evaluation under all training taxonomy classes.

\subsection{Anomaly scoring \& generative classification}  
\label{sec:anomaly-score}
When evaluating the conditional likelihood $p_\theta(\mathbf{z}|y)$ given each known semantic class, a class-conditional normalizing flow can both assign class labels and identify unlikely feature vectors via Bayes' theorem:
\begin{equation}
\label{eq:bayes}
p(y|\mathbf{z}) = \frac{p_\theta(\mathbf{z}|y)\, P(y)}{p(\mathbf{z})},
\qquad
p(\mathbf{z}) = \sum_{y} p_\theta(\mathbf{z}|y)\, P(y).
\end{equation}
We estimate the empirical class prior $P(y)$ from the training set. The posterior is used for generative classification and
the marginal likelihood serves as an anomaly score. 

Since the absolute scale of likelihoods is not comparable across the three flows,
we normalize each map by median absolute deviation normalization with statistics
computed on the validation set.
This normalization is robust to the outlier values introduced by anomalous pixels,
which would otherwise bias standard mean-variance estimates.  

Let $\mathbf{S}_l \in \mathbb{R}^{H_l \times W_l}$ be the raw marginal likelihood map,
$[\mathbf{S}_l]_{h,w} = \sum_{y} p_{\theta}(\mathbf{z}_l^{(h,w)}|y)\, P(y)$,
and let $\tilde{\mathbf{S}}_l$ denote its median-absolute-deviation-normalized
counterpart. We define the compound anomaly score $\mathbf{S}$ as the negative sum
of normalized marginal likelihoods over all three feature stages:
\begin{equation}
\mathbf{S} =
-(
\tilde{\mathbf{S}}_{\mathrm{ViT}}
+
\tilde{\mathbf{S}}_{\mathrm{res4}}
+
\tilde{\mathbf{S}}^{\uparrow2}_{\mathrm{res5}}
)
\ \in \mathbb{R}^{\frac{H}{16} \times \frac{W}{16}},
\label{eq:anomaly_score}
\end{equation}
where $^{\uparrow2}$ denotes bilinear upsampling by factor 2.


Similarly, we perform generative classification. Let $\tens{K}_l \in \mathbb{R}^{H_l \times W_l \times Y}$ be a volume of posteriors given every known class, $[\tens{K}_l]_{h,w,y} = p(y|\mathbf{z}_l^{(h,w)})$.
For every feature stage $l$, the class prediction then follows as $\hat{\mathbf{Y}}_l = \arg\max_y \tens{K}_l$ with the final semantic label obtained by majority vote.

\subsection{Segment-guided aggregation \& inlier suppression}
\label{sec:sam_refinement}

Given the anomaly score $\mathbf{S}$ (Eq.~\ref{eq:anomaly_score}),
we refine per-location predictions into spatially coherent segments with SAM3~\cite{carion2025arxiv} using two prompting modalities:
point prompts for score aggregation and text prompts for inlier suppression.

\noindent\textbf{Segment-guided score aggregation.}
The per-location independence assumption 
(Sec.~\ref{sec:normalizing_flows}) discards spatial structure, yielding 
predictions that lack coherence within object regions. 
We recover spatial consistency by prompting SAM3 with a regular point grid~\cite{ravi2024arxiv}.
We binarize the $N$ obtained class-agnostic probabilistic masks
 $\{\mathbf{M}_n\}_{n=1}^{N}$ as $\bar{\mathbf{M}}_n = \llbracket \mathbf{M}_n>0.5 \rrbracket$.
We compute the anomaly score of a mask $s_n$ as the mean of the anomaly score (Eq.~\ref{eq:anomaly_score})
over the foreground region,
and define the aggregated score map 
as a combination of probabilistic masks weighted by mask-wide scores:
\begin{equation}
    \mathbf{S}_{\mathrm{seg}} = \frac{\sum_{n} s_n\, \mathbf{M}_n}{\sum_{n} \mathbf{M}_n},
    \qquad    
    s_n = \frac{\sum_{i,j} \mathbf{S}_{i,j}\, [\bar{\mathbf{M}}_n]_{i,j}}
     {\sum_{i,j}  [\bar{\mathbf{M}}_n]_{i,j}}.
    \label{eq:aggregation}
\end{equation}
This enforces score consistency within each segment while 
handling overlapping masks.

\noindent\textbf{Text-prompted inlier suppression.}
Segment-level aggregation can introduce false positives when masks 
covering in-distribution regions falsely spread high anomaly scores 
from noisy per-location predictions. Since \method~targets road-driving 
scenes, we exploit the fixed semantic taxonomy of Cityscapes and query 
SAM3 with text prompts corresponding to known classes. 
We prompt SAM3 with the classes: \textit{"road"}, 
\textit{"sidewalk"}, \textit{"terrain"}, \textit{"vegetation"}, and \textit{"sky"}. 
Let $\bar{\mathbf{M}}_{\mathrm{sup}}$ denote the union of binary masks obtained from the 
five inlier text prompts, indicating locations to be suppressed. We 
obtain the final anomaly score $\hat{\mathbf{S}}$ by setting suppressed 
locations to the per-image minimum, leaving all other locations unchanged:
\begin{equation}
[\hat{\mathbf{S}}]_{i,j} =
\begin{cases}
\min(\mathbf{S}_{\mathrm{seg}}) & \text{if } [\bar{\mathbf{M}}_{\mathrm{sup}}]_{i,j} = 1, \\
[\mathbf{S}_{\mathrm{seg}}]_{i,j} & \text{otherwise.}
\end{cases}
\label{eq:suppression}
\end{equation}

We select these five classes as the dominant static background ('stuff') categories in road-driving scenes that span spatially coherent regions and can be reliably localized by text grounding of SAM3.
Our inlier suppression pipeline is method-agnostic and 
consistently improves prior anomaly segmentation methods without 
retraining (Sec.~\ref{sec:discussion}).
However, its inlier prompts are specific to the road-driving taxonomy and would need to be redefined for other domains.

\subsection{Instance-level anomaly detection}
\label{sec:instances}
The SAM3 masks $\{\mathbf{M}_n\}$ computed in the aggregation step directly express anomaly instances without any training modifications.
We first binarize $\hat{\mathbf{S}}$ with a threshold $\tau_\text{inst}$ selected to maximize the pixel-wise F1 score
on the validation subset. Next, we extract anomaly instance proposals by finding connected components over the thresholded binary mask.
However, a single anomaly proposal may contain several distinct objects. We therefore replace each proposal
with the SAM3 masks $\llbracket \mathbf{M}_n>0.5 \rrbracket$ already computed in Eq.~(\ref{eq:aggregation}).
We select the mask with highest IoU with respect to the proposal and IoU ${\geq}\,0.5$.
Finally, we iteratively select masks that account for
at least 10\% of the uncovered proposal area.
We define the instance-level anomaly score by mean-pooling $\hat{\mathbf{S}}$ within its mask.

\subsection{Anomaly category discovery}
\label{sec:discovery}
Anomalous instances detected by \method~may belong to unseen semantic 
categories. We formulate their consistent categorization as a clustering 
problem of instance-level descriptors into an unknown number of novel classes.
Each instance is represented by a descriptor $\mathbf{d} \in \mathbb{R}^{D}$
formed by concatenating L2-normalized mask-average-pooled backbone features
from three scales and renormalising:
\begin{equation}
  \mathbf{d} = \frac{[\hat{\mathbf{d}}_{\mathrm{res5}}; \hat{\mathbf{d}}_{\mathrm{res4}}; \hat{\mathbf{d}}_{\mathrm{vit}}]} {\left\|[\hat{\mathbf{d}}_{\mathrm{res5}}; \hat{\mathbf{d}}_{\mathrm{res4}}; \hat{\mathbf{d}}_{\mathrm{vit}}]\right\|_2},
  \quad
    \mathbf{d}_{l} = \frac{1}{|\mathcal{I}|}\sum_{\mathbf{i} \in \mathcal{I}}\mathbf{z}_{l}^{(\mathbf{i})}
  \label{eq:descriptor}
\end{equation}
where $\mathcal{I}$ is the set of locations of pixels belonging to the instance mask.
The normalization places $\mathbf{d}$ on the unit hypersphere
$\mathbb{S}^{D-1}$, making cosine similarity the natural distance metric.
We cluster $\{\mathbf{d}_i\}_{i=1}^{N_\text{inst}}$ using agglomerative clustering
with cosine distance and average linkage.
The number of clusters $K$ is not specified in advance
but emerges from a similarity threshold $\tau_\text{sim}$ as following. 
Starting from $N_\text{inst}$ singleton clusters, the algorithm repeatedly merges the
closest pair until all remaining cluster pairs have average-linkage 
 exceeding $1 - \tau_\text{sim}$.

\section{Experimental setup}
We introduce the tasks that SAFe addresses, including the datasets and evaluation protocols.
Supplementary source code~\footnote{https://anonymous.4open.science/r/SAFe-5A1B}
includes hyperparameters and implementation details.

\label{sec:experiments}
\subsection{Anomaly-aware tasks}
\method~detects anomalous segments, instances and categories while performing generative classification. 
We evaluate its performance and compare with baselines on the following  anomaly-aware tasks already established in prior work.
Anomaly segmentation \cite{chan2021neuripsdb} performs binary pixel-wise classification into in-distribution and anomalous pixels.
Anomaly instance segmentation \cite{nekrasov2025icra} extends this by additionally distinguishing individual anomalous object instances.
Open-set semantic segmentation \cite{sodano2024arxiv} assigns each pixel to a predefined semantic class or labels it as anomalous.
Open-world semantic segmentation \cite{sodano2024arxiv} goes further by discovering novel classes within anomalous regions. 
The anomaly-aware panoptic setting \cite{sodano2024arxiv} departs from standard closed-set panoptic segmentation by evaluating in the anomalous region without classifying in the inlier region. Open-set panoptic segmentation requires only class-agnostic instance IDs for objects found inside the anomalous region, with the inlier region collapsed into a single undifferentiated stuff blob for scoring purposes, while open-world panoptic segmentation additionally discovers novel semantic categories.

\subsection{Datasets and Evaluation}
We train all models on Cityscapes~\cite{cordts2016cvpr} unless otherwise specified, with the standard 19-class 
taxonomy and evaluate as described on the following road-driving benchmarks.

\noindent\textbf{SMIYC}~\cite{chan2021neuripsdb} provides two anomaly segmentation tracks: 
\textit{AnomalyTrack}, targeting larger anomalies anywhere in the scene, and 
\textit{ObstacleTrack}, focusing on small obstacles on the road surface.
We report pixel-level metrics average precision (AP) and false positive rate at 95\% true positive rate 
(FPR$_{95}$).

\noindent\textbf{OoDIS}~\cite{nekrasov2025icra} extends SMIYC and Fishyscapes 
L\&F~\cite{blum2021ijcv} with instance-level labels, allowing evaluation of anomaly 
instance segmentation.
We report instance-level AP and AP at an IoU threshold of 50\% (AP$_{50}$).

\noindent\textbf{ISSU}~\cite{laskar2025cvpr} is a large-scale benchmark with a dedicated 
training set, offering both in-domain and cross-domain evaluation splits under the 
ISSU and Cityscapes training datasets, respectively.
The test set contains both closed-set and anomaly labels, enabling open-set semantic segmentation evaluation.
Beyond AP, FPR$_{95}$ and the complementary TPR$_{5}$ for anomaly segmentation, we report open IoU~\cite{grcic2024tpami} at 95\% true positive rate 
($\mathrm{oIoU}_{95}$) and at 5\% false positive rate ($\mathrm{oIoU}_{5}$) for open-set segmentation.

\noindent\textbf{PANIC}~\cite{sodano2024arxiv} provides pixel-wise semantic, instance, 
and category annotations of anomalies.
We report AP and FPR$_{95}$ for anomaly segmentation.
For open-world semantic segmentation, we report mean IoU over discovered unknown 
categories ($\overline{\text{IoU}}$), alongside homogeneity (Hom) and completeness (Com) \cite{sodano2024arxiv}.
For open-set panoptic segmentation, we report PQ, SQ, and RQ for \textit{known} stuff 
and \textit{unknown} instances.
For open-world panoptic segmentation, PQ, SQ, and RQ are decomposed across multiple 
discovered categories.


\section{Experimental results}

We evaluate \method~against prior methods on the PANIC \cite{sodano2024arxiv}, OoDIS \cite{nekrasov2025icra}, ISSU \cite{laskar2025cvpr} and SMIYC \cite{chan2021neuripsdb} benchmarks. 

Table \ref{tab:panic_segmentation} reports the performance of anomaly and open-world segmentation on PANIC. \method~achieves the lowest FPR$_{95}$ with a reduction of 11.9 pp. 
Under open-world evaluation, \method~achieves the highest homogeneity score with a gain of 14.1 pp over the best baseline. 
Baselines rely on distinct, specialized procedures for open-world semantic and panoptic tasks, whereas SAFe solves both through a single, unified instance-level approach which trades some pixel-level accuracy for multi-task consistency.

\begin{table}[H]
\centering
\footnotesize
\setlength{\tabcolsep}{4pt}
\caption{Anomaly \& open-world segmentation on PANIC.}
\label{tab:panic_semantic}
\begin{tabular}{lccccc}
\toprule
& \multicolumn{2}{c}{binary}
& \multicolumn{3}{c}{open-world} \\
\cmidrule(r){2-3}
\cmidrule(l){4-6}
Method
& AP$\,\uparrow$ & FPR$_\text{95}\downarrow$
& $\overline{\text{IoU}}\uparrow$ & Com.$\,\uparrow$ & Hom.$\,\uparrow$ \\
\midrule
ContMAV \cite{sodano2024cvpr}
& 91.7 & 66.4
& 15.8 & \textbf{81.4} & 77.3 \\

Con2MAV\cite{sodano2024arxiv}
& \textbf{95.7} & 35.3
& \textbf{20.2} & 77.6 & 78.3 \\

\method
& 94.9 & \textbf{23.4}
& 15.2 & 71.8 & \textbf{92.4} \\
\bottomrule
\end{tabular}
\label{tab:panic_segmentation}
\end{table}

\method~achieves the state-of-the-art in terms of all metrics for both panoptic tasks on PANIC (Table \ref{tab:panic_panoptic}) and for instance segmentation on OoDIS (Table \ref{tab:smiyc2}).

\begin{table}[H]
\centering
\footnotesize
\setlength{\tabcolsep}{4pt}
\caption{Anomaly-aware panoptic performance on PANIC.}
\label{tab:panic_panoptic}
\begin{tabular}{lcccccc}
\toprule
& \multicolumn{3}{c}{open-set}
& \multicolumn{3}{c}{open-world} \\
\cmidrule(r){2-4}
\cmidrule(l){5-7}
Method
& PQ$\,\uparrow$ & SQ$\,\uparrow$ & RQ$\,\uparrow$
& PQ$\,\uparrow$ & SQ$\,\uparrow$ & RQ$\,\uparrow$ \\
\midrule

Con2MAV\cite{sodano2024arxiv}
& 21.6 & 72.4 & 28.4
& 25.1 & 46.8 & 28.8 \\

Prior2Former \cite{schmidt2025iccv}
& 52.9 & 87.1 & 56.7
& - & - & - \\

\method
& \textbf{62.0} & \textbf{90.1} & \textbf{64.5}
& \textbf{50.3} & \textbf{66.0} & \textbf{54.2} \\
\bottomrule
\end{tabular}
\label{tab:panic_panoptic}
\end{table}

\begin{table}[!ht]
\centering
\footnotesize
\setlength{\tabcolsep}{4pt}
\caption{Anomaly instance segmentation on OoDIS.}
\label{tab:smiyc2}
\begin{tabular}{lcccccc}
\toprule
\multicolumn{1}{c}{} & \multicolumn{2}{c}{FS Lost\&Found} & \multicolumn{2}{c}{Road Anomaly} & \multicolumn{2}{c}{Road Obstacle}\\
Method & AP$\,\uparrow$ & AP50$\,\uparrow$ & AP$\,\uparrow$ & AP50$\,\uparrow$ & AP$\,\uparrow$ & AP50$\,\uparrow$  \\
\midrule
M2A \cite{rai2023iccv} & 11.7 & 23.6 & 4.8 & 9.0 & 17.2 & 28.4  \\
UGainS \cite{nekrasov2023gcpr}  & 27.1 & 45.8 & 11.4 & 19.2 & 27.2 & 46.5  \\
\method~& \textbf{32.6} & \textbf{51.6} & \textbf{17.7} & \textbf{26.3} & \textbf{41.7} & \textbf{66.7} \\
\midrule
\end{tabular}
\end{table}

Table~\ref{tab:issu} reports results on the ISSU 
benchmark for anomaly segmentation and open-set classification. 
\method~performs best on Road Obstacle on both temporal and static subsets and on both in-domain and cross-domain training setups, except for the in-domain static where it achieves second best FPR$_{95}$.
Results indicate that generative classification with SAFe provides a viable alterative to discriminative approaches for open-set segmentation. We classify following the generative procedure introduced in Sec.~\ref{sec:anomaly-score}, with ties broken by empirically selected $\mathbf{k}_\text{res4}$.

\begin{table*}[!ht]
\vspace{5pt}
\caption{
Anomaly segmentation and open-set segmentation performance on the ISSU benchmark.
}
\vspace{-2pt}
\label{tab:issu}
\centering
\footnotesize
\setlength{\tabcolsep}{1.5pt}
\begin{tabular}{@{}c@{\hspace{4pt}}|lcccc|ccccc|ccccc@{}}
\toprule
{} & {} & \multicolumn{4}{c|}{Road Obstacle} & \multicolumn{10}{c}{Road Anomaly}\\
\cmidrule(lr){3-6}\cmidrule(lr){7-16}
{} & {} & \multicolumn{2}{c}{Temporal} & \multicolumn{2}{c|}{Static} &
\multicolumn{5}{c|}{Temporal} & \multicolumn{5}{c}{Static}\\
\cmidrule(lr){3-4}
\cmidrule(lr){5-6}
\cmidrule(lr){7-11}
\cmidrule(lr){12-16}
{} & Method
& AP$\,\uparrow$ & $\mathrm{FPR}_{95}\downarrow$
& AP$\,\uparrow$ & $\mathrm{FPR}_{95}\downarrow$
& AP$\,\uparrow$ & $\mathrm{FPR}_{95}\downarrow$ & mIoU$\,\uparrow$ & $\mathrm{oIoU}_{95}\uparrow$ & $\mathrm{oIoU}_{5}\uparrow$
& AP$\,\uparrow$ & $\mathrm{FPR}_{95}\downarrow$ & mIoU$\,\uparrow$ & $\mathrm{oIoU}_{95}\uparrow$ & $\mathrm{oIoU}_{5}\uparrow$\\
\midrule
\multirow{7}{*}{\rotatebox[origin=c]{90}{In-domain}}
& M2A \cite{rai2023iccv} & 30.0 & 79.5 & 48.9 & 78.5 &10.7&78.6&40.3&16.9&34.1&32.0&66.9&53.9&31.5&49.5\\
& Pebal \cite{tian2022eccv} & 48.9 & 23.8 & 92.5 & 1.9 &23.6&\underline{24.7}&57.8&\textbf{46.2}&\textbf{55.7}&64.5&4.4&72.9&\underline{67.8}&\underline{67.1}\\
& UNO \cite{delic2024bmvc} & 56.1 & 92.3 & 94.0 & \textbf{1.2} &30.4&89.7&\underline{59.0}&9.8&55.2&71.4&\textbf{3.0}&\underline{73.7}&\textbf{68.4}&65.9\\
& RbA \cite{nayal2023iccv} & 57.2 & 33.7 &\textbf{95.8} & 1.7 &37.7&29.4&57.8&\underline{41.7}&\underline{55.6}&\textbf{79.1}&\underline{3.9}&72.9&\underline{67.8}&66.5\\
& EAM \cite{grcic2023cvprw} & 62.1 & 96.2 & \underline{95.6} & 1.6 &\underline{38.7}&91.4&\textbf{59.5}&6.7&\textbf{55.7}&\underline{76.8}&4.2&\textbf{73.8}&\textbf{68.4}&\textbf{67.6}\\
& PixOOD \cite{vojivr2024eccv} & \underline{83.1} & \underline{10.1} & 93.1 & 4.3 &6.2&56.5&55.1&32.9&52.2&20.3&39.4&65.8&47.8&60.9\\
& \method~& \textbf{87.4} & \textbf{4.4} & \textbf{95.8} & \underline{1.5} &\textbf{54.2}&\textbf{19.0}&56.7&17.3&37.8&58.8&17.7&64.5&23.7&45.6\\
\midrule
\multirow{7}{*}{\rotatebox[origin=c]{90}{Cross-domain}}
& M2A \cite{rai2023iccv} & 30.7 & 81.3 & 63.3 & 45.8 &10.7&91.9&34.0&\textbf{16.8}&33.3&37.5&79.8&50.6&\underline{26.4}&48.2\\
& Pebal\cite{tian2022eccv} & 29.4 & 67.1 & 73.6 & 40.8 &16.1&\underline{79.5}&43.7&8.3&42.1&48.3&\underline{64.9}&57.5&\textbf{34.2}&55.4\\
& UNO \cite{delic2024bmvc} & 49.1 & 90.5 & 66.3 & 90.8 &\textbf{37.2}&92.4&\textbf{57.4}&6.6&\textbf{54.6}&\underline{55.5}&92.9&\textbf{68.1}&12.0&\textbf{65.6}\\
& RbA \cite{nayal2023iccv} & 37.9 & 87.9 & 76.1 & 68.9 &24.6&91.6&43.7&3.2&41.9&\textbf{56.4}&80.7&57.5&11.9&55.1\\
& EAM \cite{grcic2023cvprw} & 43.0 & 98.3 & 61.4 & 93.4 &\underline{35.6}&96.4&\underline{57.3}&2.7&\underline{53.2}&54.5&95.4&\underline{66.8}&7.9&\underline{63.4}\\
& PixOOD \cite{vojivr2024eccv} & \underline{84.3} & \underline{10.8} & \underline{92.3} & \underline{5.1} &4.8&80.7&48.7&14.7&47.0&11.4&73.7&56.3&20.4&52.8\\
& \method~& \textbf{85.8} & \textbf{4.8} & \textbf{93.8} & \textbf{1.7} &28.7&\textbf{20.4}&46.1&\underline{15.0}&33.1&32.1&\textbf{16.9}&52.7&24.5&41.9\\
\bottomrule
\end{tabular}
\end{table*}

Table~\ref{tab:smiyc} presents SMIYC benchmark results.
Unlike the bottom section, the top section methods use foundation models.
\method~attains state-of-the-art on ObstacleTrack and accomplishes best AP on AnomalyTrack when considering related methods that build on foundation models.

\begin{table}[h]
\caption{Anomaly segmentation performance on SMIYC.}
\label{tab:smiyc}
\centering
\footnotesize
\setlength{\tabcolsep}{8pt}
\begin{tabular}{lcccc@{}}
\toprule
 & \multicolumn{2}{c}{ObstacleTrack} & \multicolumn{2}{c}{AnomalyTrack}\\
\cmidrule(lr){2-3}
\cmidrule(lr){4-5}
Method &
AP$\,\uparrow$ & FPR$_{95}\downarrow$
& AP$\,\uparrow$ & FPR$_{95}\downarrow$ \\
\midrule
EAM \cite{grcic2023cvprw} &92.9&0.5&93.8&4.1 \\
UNO \cite{delic2024bmvc} &93.2&0.2&\textbf{96.3}&\textbf{2.0} \\
RbA \cite{nayal2023iccv} &\underline{95.1}&0.1&94.5&4.6 \\
\midrule
VL4AD  \cite{zhong2024vl4ad} &78.7&0.6&92.9&\underline{3.3} \\
PixOOD \cite{vojivr2024eccv} &88.9&0.3&68.9&54.3 \\
ULRE\cite{holle2025iccv} &91.1&0.2&94.2&5.8 \\
SOTA \cite{zheng2025icm} &91.9&0.1&93.9&7.9 \\
FlowCLAS  \cite{lee2026wacv} &94.2&0.1&94.3&6.6 \\
UEM  \cite{nayal2025ijcv} &94.4&0.1&95.6&4.7 \\

\method~&\textbf{95.5}&\textbf{0.04}&\underline{95.9}& 4.1 \\
\bottomrule
\end{tabular}
\end{table}

\section{Discussion}
\label{sec:discussion}
We analyze the key design choices of \method~through targeted ablations presented in Table~\ref{tab:ablation_all}. We report AP and FPR$_{95}$ averaged over the validation datasets RoadAnomaly \cite{lis2019iccv} and Fishyscapes (FS) Static and Lost\&Found \cite{blum2021ijcv}.

\subsection{On backbone architecture and training}
Table~\ref{tab:ablation_backbone} compares convolutional and transformer architectures under cross-entropy ImageNet supervision (CE-IN) versus DINO self-supervision, using a single normalizing flow and its likelihood as anomaly score.
Self-supervision with DINOv3 attains the best result in both groups, and matters more than architecture or capacity. DINOv3 ViT-L surpasses its DINOv2 counterpart by 3.2~pp AP and 5.4~pp FPR$_{95}$, while ConvNeXt-L outperforms the smaller DINOv3-trained ConvNeXt-B only marginally (0.5~pp AP, 0.8~pp FPR$_{95}$). We use ConvNeXt-L and Swin-L~\cite{liu2021iccv} as CE-IN baselines, with the latter as an alternative to ViT given its establishment as a segmentation backbone~\cite{cheng2022cvpr}. DINOv3 pre-trained backbones surpass their CE-IN counterparts by up to 49.9~pp AP and 33.2~pp FPR$_{95}$.
In this single-level setup, DINO-pretrained transformers outperform their convolutional counterparts. However, this gap narrows substantially in the full \method~pipeline, as further discussed.

\subsection{On using multiple backbones and feature stages}
Table~\ref{tab:ablation_feature_level} evaluates the impact of different feature stages, precisely L$_{17}$ and L$_{23}$ from ViT-L and res4 and res5 from ConvNeXt-L. We score features at each stage by a separate normalizing flow and fuse them into the final anomaly score.
Results reveal that using only the final level of ViT but both res4 and res5 of ConvNeXt gives best performance.
We attribute this to receptive field of ConvNeXt, which grows with depth and makes res4 and res5 complementary in spatial context. In contrast, the self-attention of ViT, is globally contextualized from the outset, making features across depth more redundant.
Backbone fusion further improves results. 
However, including the intermediate stage of ViT worsens FPR$_{95}$.
Thus, we keep L$_{23}$, res4 and res5 in our final configuration and train three normalizing flows.

\subsection{On normalizing flow architecture}
In Table~\ref{tab:nf_architecture}
we ablate the depth, width (upper) and gating activation \cite{shazeer2020arxiv} (lower) inside the conditioner of the normalizing flow.
As empirically indicated, we adopt 16 steps with 2 blocks each as our final configuration. 
We show that gated mechanisms~\cite{shazeer2020arxiv} outperform the standard MLP architecture. 
We use SwiGLU for gating, as it achieves a lower FPR$_{95}$ than the GEGLU counterpart, and the softer negative lobe of Swish better preserves the class-conditional signal introduced by FiLM modulation. 

\subsection{On class-conditional density modeling}
\label{sec:discussion-class-cond}
Table~\ref{tab:nf_conditioning} ablates three axes of class-conditioning: embedding concatenation (C), FiLM modulation (F), and a learned conditional base distribution (B).
Each axis independently improves over the unconditional baseline, and combining all three attains the best result, a 10.7 pp absolute AP improvement and 1 pp lower FPR$_{95}$.
The lower part isolates two special cases. Fixing the base distribution to orthogonal means with unit variance degrades performance, confirming that learning the Gaussian parameters is necessary to capture class-specific density accurately. Replacing the jointly optimized embedding with a fixed one-hot encoding degrades performance substantially, showing that the embedding must be learned to capture meaningful inter-class relationships.

\begin{table*}[!ht]
\centering
\footnotesize
\vspace{4pt}
\caption{\textbf{A.} DINOv3 self-supervision outperforms ImageNet cross-entropy pretraining (CE-IN). \textbf{B.} Feature-level fusion within and across backbones. \textbf{C.} Normalizing-flow depth, width, and gating. \textbf{D.} Class-conditioning axes: embedding concatenation (C), FiLM (F), conditional base distribution (B). 
}
\begin{subtable}{0.26\linewidth}
\centering
\setlength{\tabcolsep}{3pt}

\caption{Backbone \& pre-training}
\begin{tabular}{l l c c}
\toprule
Training & Arch. & AP & FPR$_{95}$ \\
\midrule
CE-IN  & ConvNeXt-L & 51.9 & 17.7 \\
DINOv3 & ConvNeXt-B & 68.2 & 8.6 \\
\textbf{DINOv3} & \textbf{ConvNeXt-L} & \textbf{68.7} & \textbf{7.8} \\
\midrule
CE-IN  & Swin-L & 27.6 & 36.6 \\
DINOv2 & ViT-L  & 74.3 & 8.8 \\
\textbf{DINOv3} & \textbf{ViT-L}  & \textbf{77.5} & \textbf{3.4} \\
\bottomrule
\end{tabular}
\label{tab:ablation_backbone}
\end{subtable}
\hfill
\begin{subtable}{0.24\linewidth}
\centering
\setlength{\tabcolsep}{3pt}
\caption{Feature-level fusion}
\begin{tabular}{lcc}
\toprule
Feature levels & AP & FPR$_{95}$ \\
\midrule
$\mathrm{L}_{17}$ & 65.1 & 7.9  \\
$\mathrm{L}_{23}$ & 77.5 & \textbf{3.4} \\
$\mathrm{L}_{17}$, $\mathrm{L}_{23}$ & 77.2 & 3.9 \\
\midrule
res4 & 66.8 & 10.2 \\
res5 & 68.7 & 7.8 \\
res4, res5 & 75.9 & 4.8 \\
\midrule
$\mathrm{L}_{23}$, res4, res5 & \textbf{82.5} & \textbf{3.4}  \\
$\mathbf{L}_{17}$, $\mathbf{L}_{23}$, \textbf{res4}, \textbf{res5} & \textbf{82.5} & 3.7  \\
\bottomrule
\end{tabular}
\label{tab:ablation_feature_level}
\end{subtable}
\hfill
\begin{subtable}{0.2\linewidth}
\centering
\setlength{\tabcolsep}{3pt}
\caption{Flow design}
\begin{tabular}{c c c c}
\toprule
Steps & Blocks & AP & FPR$_{95}$ \\
\midrule
12 & 2 & 66.5 & 8.2 \\
\textbf{16} & \textbf{2} & \textbf{68.7} & \textbf{7.8} \\
16 & 1 & 65.7 & 8.6 \\
\bottomrule\\
[-6pt]
\multicolumn{2}{l}{Layer} & AP & FPR$_{95}$ \\
\midrule
\multicolumn{2}{l}{MLP}    & 64.7 & 8.8 \\
\multicolumn{2}{l}{ReGLU}  & 66.8 & 8.2 \\
\multicolumn{2}{l}{GEGLU}  & \textbf{69.1 }& 8.1 \\
\multicolumn{2}{l}{\textbf{SwiGLU}} & 68.7 & \textbf{7.8} \\
\bottomrule
\end{tabular}
\label{tab:nf_architecture}
\end{subtable}
\hfill
\begin{subtable}{0.23\linewidth}
\centering
\setlength{\tabcolsep}{3pt}
\caption{Class-conditioning}
\begin{tabular}{ccc c c}
\toprule
C & F & B & AP & FPR$_{95}$ \\
\midrule
\xmark & \xmark & \xmark & 58.0 & 8.8 \\
\cmark & \xmark & \cmark & 67.4 & 8.4 \\
\xmark & \cmark & \cmark & 67.5 & 8.1 \\
\cmark & \cmark & \xmark & 68.1 & 8.2 \\
\cmark & \cmark & \cmark & \textbf{68.7} & \textbf{7.8} \\
\midrule
\cmark & \cmark & \scriptsize{fixed} & 67.9 & 7.9 \\
\scriptsize{OH} & \scriptsize{OH} & \cmark & 64.9 & 8.1 \\
\bottomrule
\end{tabular}
\label{tab:nf_conditioning}
\end{subtable}
\label{tab:ablation_all}
\end{table*}

\subsection{On SAM3 segment aggregation and inlier suppression}

Table~\ref{tab:sam3_ablation} measures separate contributions of segment-guided score aggregation (A) and text-prompted inlier suppression (S).
We show improvements to~\method~and two representative baselines, namely UNO and PixOOD.
Segment-guided score aggregation alone improves AP in most settings, with two eceptions: UNO on RoadAnomaly (-0.4 pp) and PixOOD on FS Static (-2.5 pp).
Aggregation can worsen FPR$_{95}$ by increasing scores over false-positive segments.
Text-prompted inlier suppression corrects this, additionally boosting AP while consistently improving FPR$_{95}$.
For~\method, combining both stages yields the best overall result, with one exception: FPR95 on FS L\&F increases, a trade-off offset by a 20.9 pp gain in AP.
Overall, aggregation and suppression provide complementary benefits that neither achieves alone.

\begin{table}[H]
\centering
\footnotesize
\caption{%
  Impact of SAM3-based score aggregation~(A) and inlier suppression~(S) on \method~and other baselines.
}
\setlength{\tabcolsep}{4pt}
\begin{tabular}{l cc cccccc}
\toprule
 
\multirow{3}{*}{Method}
  & \multirow{3}{*}{A}
  & \multirow{3}{*}{S}
  & \multicolumn{2}{c}{RoadAnomaly}
  & \multicolumn{2}{c}{FS Static}
  & \multicolumn{2}{c}{FS L\&F} \\
& & &
{AP} & {FPR$_{95}$} &
{AP} & {FPR$_{95}$} &
{AP} & {FPR$_{95}$} \\

\midrule

\multirow{3}{*}{UNO~\cite{delic2024bmvc}}
  & \xmark & \xmark
  & 88.5 
  & \textbf{7.4 } 
  & 98.0 
  & 0.04  
  & 81.8 
  & 1.3    \\
 
  & \cmark & \xmark
  & 88.1 
  & 9.1  
  & \textbf{99.4} 
  & \textbf{0.01} 
  & 88.7 
  & 2.7   \\
 
  & \cmark & \cmark
  & \textbf{88.6} 
  & 7.6  
  & \textbf{99.4} 
  & \textbf{0.01} 
  & \textbf{89.1} 
  & \textbf{1.1} \\
 
\midrule

\multirow{3}{*}{PixOOD~\cite{vojivr2024eccv}}
  & \xmark & \xmark
  & 66.4
  & 27.9
  & \textbf{96.6} 
  & \textbf{0.2}  
  & 40.6 
  & 8.9   \\
 
  & \cmark & \xmark
  & 72.3 
  & 25.1
  & 94.1
  & 1.3  
  & 66.5 
  & 13.0  \\
 
  & \cmark & \cmark
  & \textbf{77.8} 
  & \textbf{14.9} 
  & 95.9 
  & 0.5  
  & \textbf{73.1 }
  & \textbf{5.2}  \\
 
\midrule
 
\multirow{3}{*}{\method}
  & \xmark & \xmark
  & 89.7
  & 7.6 
  & 97.5 
  & 0.2 
  & 60.4 
  & \textbf{2.5}  \\
 
  & \cmark & \xmark
  & 92.7 
  & 7.7 
  & \textbf{99.6} 
  & \textbf{0.004} 
  & 77.2 
  & 6.7  \\
 
  & \cmark & \cmark
  & \textbf{93.7} 
  & \textbf{5.8} 
  & \textbf{99.6} 
  & \textbf{0.004} 
  & \textbf{81.3} 
  & 11.0          \\
 
\midrule
\end{tabular}
\label{tab:sam3_ablation}
\end{table}



\subsection{On clustering algorithms}

Table~\ref{tab:clustering} compares the used similarity-based agglomerative clustering
against representative strategies on the PANIC validation split: spherical $k$-means and spectral clustering with $K$ selected by silhouette score,
GMM and movMF~\cite{banerjee2005jmlr} with $K$ selected by the Bayesian Information Criterion (BIC), DBSCAN and HDBSCAN~\cite{campello2013density} with implicit selection of $K$, and agglomerative clustering.
We select method-specific hyperparameters via grid search.
Despite offering automatic K selection, density-based methods proved poorly suited to our descriptor space: DBSCAN collapsed into a single cluster, while HDBSCAN over-segmented. 
Spectral clustering achieved the highest completeness without collapsing
Agglomerative clustering with a similarity-based stopping criterion performed best on our normalized, cosine-structured descriptors, outperforming BIC- and silhouette-based selection of K. We present this as a supporting case study for our specific setting rather than a general comparison of clustering algorithms as other hyperparameter choices or descriptor spaces may favor different methods.

\begin{table}[H]
\centering
\caption{Comparison of clustering algorithms for anomaly category discovery
on the PANIC validation split.}
\label{tab:clustering}
\setlength{\tabcolsep}{4pt}
\begin{tabular}{l cc ccc}
\toprule
Method  & Compl. & Hom. & PQ & SQ & RQ \\
\midrule
DBSCAN  &100.0 & 27.0 &  0.0 &  3.0 &  0.0 \\
HDBSCAN  & 81.8 & 94.8 &  6.3 & 11.4 &  7.0 \\
GMM  & 77.9 & 79.5 & 11.0 & 25.6 & 12.3 \\
movMF  & 78.0 & 81.6 & 9.6 & 22.6 & 10.8 \\
Spherical $k$-means  & 74.0 &	78.0 &	9.2 &	20.3 &	10.3 \\
Spectral  & \textbf{85.3} & 88.7 & 13.8 & 26.4 & 15.5 \\
\midrule
Agglomerative-silhouette  & 79.1 & 92.7 & 11.9 & 23.1 & 13.5 \\
Agglomerative-BIC  & 78.3 & 93.4 & 12.8 & 21.2 & 12.6 \\
Agglomerative-similarity & 74.6 & \textbf{97.7} & \textbf{24.5} & \textbf{42.5} & \textbf{28.9} \\
\bottomrule
\end{tabular}
\end{table}

\subsection{On the computational cost}
The normalizing flows have 258M trainable parameters in total, with peak GPU memory usage of 4.8 GB at batch size 2048. As the backbones and SAM3 remain frozen throughout, this constitutes the entire training cost of \method.
At inference, SAFE requires 7.7 s per image at 1024×2048 resolution, where the SAM3-guided mask refinement dominates this cost, accounting for 75\%.
Experiments were conducted on Nvidia RTX A6000 GPU.

\section{Conclusion}
\label{sec:conclusion}
We introduced~\method, an anomaly-aware segmentation method that extends to instance detection and concept grouping within anomalous regions.
The method performs per-class density estimation over frozen foundation features without fine-tuning visual backbones.
\method~jointly trains class-conditional normalizing flows over transformer and convolutional representations, and we empirically demonstrate their complementary interaction.
The resulting density estimator additionally supports generative classification without architectural modifications.
We also proposed a post hoc step that aggregates per-location anomaly scores into coherent segments using SAM3.
This refinement also improves existing anomaly segmentation methods, and enables instance-level recognition without a dedicated instance head.
The current inlier suppression is tied to the road-driving taxonomy, limiting direct transfer to other application domains.
Therefore, promising future directions include applications beyond road-driving scenes, temporal consistency in video-based anomaly scoring, and downstream control-loop integration.

\section{Acknowledgments}
This research has been funded by the H2020 project AIFORS under Grant Agreement No 952275 and supported by the Croatian Science Foundation
under contract No DOK-2025-02-5149.

\bibliographystyle{IEEEtran}
\bibliography{references}

\end{document}